\documentclass{article}
\usepackage{iclr2022_conference}
\iclrfinalcopy

\usepackage{times}
\usepackage{graphicx}
\usepackage{amsthm}
\newtheorem{theorem}{Theorem}[section]
\newtheorem{lemma}[theorem]{Lemma}

\usepackage{amsmath,amsfonts,bm}

\def\eqref#1{equation~\ref{#1}}

\def\1{\bm{1}}

\def\ry{{\textnormal{y}}}

\def\rvtheta{{\mathbf{\theta}}}

\def\rvx{{\mathbf{x}}}

\def\vtheta{{\bm{\theta}}}

\def\vx{{\bm{x}}}

\DeclareMathAlphabet{\mathsfit}{\encodingdefault}{\sfdefault}{m}{sl}
\SetMathAlphabet{\mathsfit}{bold}{\encodingdefault}{\sfdefault}{bx}{n}

\def\sX{{\mathbb{X}}}
\def\sY{{\mathbb{Y}}}

\newcommand{\R}{\mathbb{R}}

\usepackage{hyperref}
\usepackage{url}

\usepackage{multirow}
\title{Universally rank consistent ordinal regression in neural networks }

\author{Garrett Jenkinson\textsuperscript{\rm $\dagger$,2,3}, Gavin R. Oliver\textsuperscript{\rm $\dagger$,2,3}, Kia Khezeli\textsuperscript{\rm 2,3},  John Kalantari \textsuperscript{\rm 2,3,5} \& Eric W. Klee \textsuperscript{\rm 1,2,3,4}\\
\texttt{\scriptsize \{Jenkinson.William,Oliver.Gavin,Khezeli.Kia,Kalantari.John,Klee.Eric\}@mayo.edu}\\ 
\textsuperscript{\rm 1}Department of Laboratory Medicine and Pathology, Mayo Clinic, Rochester, MN, USA\\
\textsuperscript{\rm 2}Department of Quantitative Health Sciences, Mayo Clinic, Rochester, MN, USA\\
\textsuperscript{\rm 3}Center for Individualized Medicine, Mayo Clinic, Rochester, MN, USA\\
\textsuperscript{\rm 4}Department of Clinical Genomics, Mayo Clinic, Rochester, MN, USA\\
\textsuperscript{\rm 5}Department of Surgery, Mayo Clinic, Rochester, MN, USA\\
\textsuperscript{\rm $\dagger$}Equal contributors.\\
}

\begin{document}

\maketitle

\begin{abstract}
Despite the pervasiveness of ordinal labels in supervised learning, it remains common practice in deep learning to treat such problems as categorical classification using the categorical cross entropy loss. Recent methods attempting to address this issue while respecting the ordinal structure of the labels have resorted to converting ordinal regression  into a series of extended binary classification subtasks. However, the adoption of such methods remains inconsistent due to theoretical and practical limitations. Here we address these limitations by demonstrating that the subtask probabilities form a Markov chain. We show how to straightforwardly modify neural network architectures to exploit this fact and thereby constrain predictions to be universally rank consistent. We furthermore prove that all rank consistent solutions can be represented within this formulation, and derive a loss function producing maximum likelihood parameter estimates. Using diverse benchmarks and the real-world application of a specialized recurrent neural network for COVID-19 prognosis, we demonstrate the practical superiority of this method versus the current state-of-the-art. The method is open sourced as user-friendly PyTorch\footnote{\url{https://github.com/GarrettJenkinson/condor_pytorch}} and TensorFlow\footnote{\url{https://github.com/GarrettJenkinson/condor_tensorflow}} packages.
\end{abstract}

\section{Introduction}
Ordinal regression (sometimes called ordinal classification) is applied to data in which the features of the $n$-th example  $\vx_n\in\sX$ correspond to a label $y_n\in\sY:= \{r_1,\ldots,r_K\}$ from a set of elements that have a well-defined ranking or ordering $r_1<r_2<\cdots <r_K$. However, unlike traditional metric regression, the ranks cannot be assumed to have quantitative differences or distances amongst themselves. For example, while a syntactic statement such as ``terrible''$<$``great''$<$``best'' may be intuitive, it conveys nothing about quantitative distance between the ranks nor qualitative differences i.e. whether the distance between ``terrible'' and ``great'' is equal to that between ``great'' and ``best''. The aim in this setting is to build a reliable rule or regression function $h: \sX \rightarrow \sY$ from the domain of the features $\sX$ to the range of the ordinal labels $\sY$.

In the published literature for applied problems, it remains commonplace to ignore the ordering of the labels and apply categorical algorithms to such data~\citep{Levi15,Roth15}, which in neural networks often results in application of the categorical cross entropy (CCE) loss. Problematically, a categorical loss assumes all mislabeling by $h$ is equally wrong, whereas it is clear that predicting ``great'' when the true label is ``best'' would be preferable to a prediction of ``terrible''. Although the problematic nature of this practice has been recognized for more than 35 years~\citep{Forr86}, it still remains common to make the implicit or explicit assumption that ordinal data or labels exist on an interval or ratio scale. 

Most contemporary algorithms found in the ordinal regression literature~\citep{McCullagh80,Ober99,Cram02,Shas02,Raja03,Shen05,Chu05,LiLin07,Bacc09,Niu16,FernCard18,CORAL}, can be viewed through the lens of a general framework proposed by \citet{LiLin07} wherein the labels are encoded as binary vectors by an invertible encoder $e:\sY \rightarrow \{0,1\}^{K-1}$ and the regression function $h$ becomes a collection of $K-1$ binary classifiers along with the decoder $e^{-1}: \{0,1\}^{K-1} \rightarrow \sY$. However, many of these existing algorithms share one major limitation: rank-inconsistency among predictions. Briefly, the $K-1$ binary tasks are not independent and training $K-1$ classifiers that incorrectly treat them as independent will produce conflicting predictions on the binary tasks, impeding both performance and interpretation of the results. Most recently, \citet{CORAL} attempted to address this problem in deep neural network (DNN) architectures by proposing a final layer that shares weights among $K-1$ binary outputs (differing only in their bias terms). 

Herein, we identify the theoretical and practical limitations of the CORAL approach by \citet{CORAL}, and address these concerns by implementing a new algorithm `Conditionals for Ordinal Regression' (CONDOR). We prove that CONDOR is universally rank consistent and that it is sufficiently expressive to reach any rank consistent solution. In theory, the method is compatible with any combination of binary classification algorithms producing probabilities, but herein we focus on DNN architectures trained by backpropagation. Using our open source  PyTorch and TensorFlow packages, CONDOR can be implemented with minor modifications to any categorical DNN model using these frameworks, allowing for increased adoption of ordinal regression in the applied literature.

\section{Methods} 
We begin by introducing the CONDOR framework and its notations, and then provide proofs of the universal rank consistency and the full expressiveness of the framework. 

\subsection{CONDOR framework}
The \citet{LiLin07} encoder $e$ converts the ordinal regression label $y$ into $K-1$ binary classification labels $y^{(1)},\ldots,y^{(K-1)}$ using indicator variables $y^{(1)}:=\displaystyle \1_\mathrm{y > r_1}, \ldots, y^{(K-1)}:=\displaystyle \1_\mathrm{y>r_{K-1}}$, where the indicator variable $\1_\mathrm{expr}$ is defined as \begin{align*}
    \1_\mathrm{expr} := \begin{cases}
    1,& \mathrm{expr~is~true},\\
    0,& \mathrm{expr~is~false}.
    \end{cases}
\end{align*} The
ordinal classification problem then becomes a matter of producing $K-1$ binary classifier subtasks $f_k:\sX \rightarrow \{0,1\}$, which we assume come from thresholding predicted binary class probabilities $p_k:\sX\rightarrow [0,1]$ as $f_k(\vx) = \displaystyle \1_\mathrm{p_{k}(\vx)>0.5 }.$ When using parametric 
techniques such as DNNs, these functions will be parameterized
by $\vtheta\in\Theta$ and denoted notationally as $p_k(\vx;\vtheta)$
and $f_k(\vx;\vtheta)$.

\begin{figure}[ht]
\begin{center}
\includegraphics[width=0.6\linewidth]{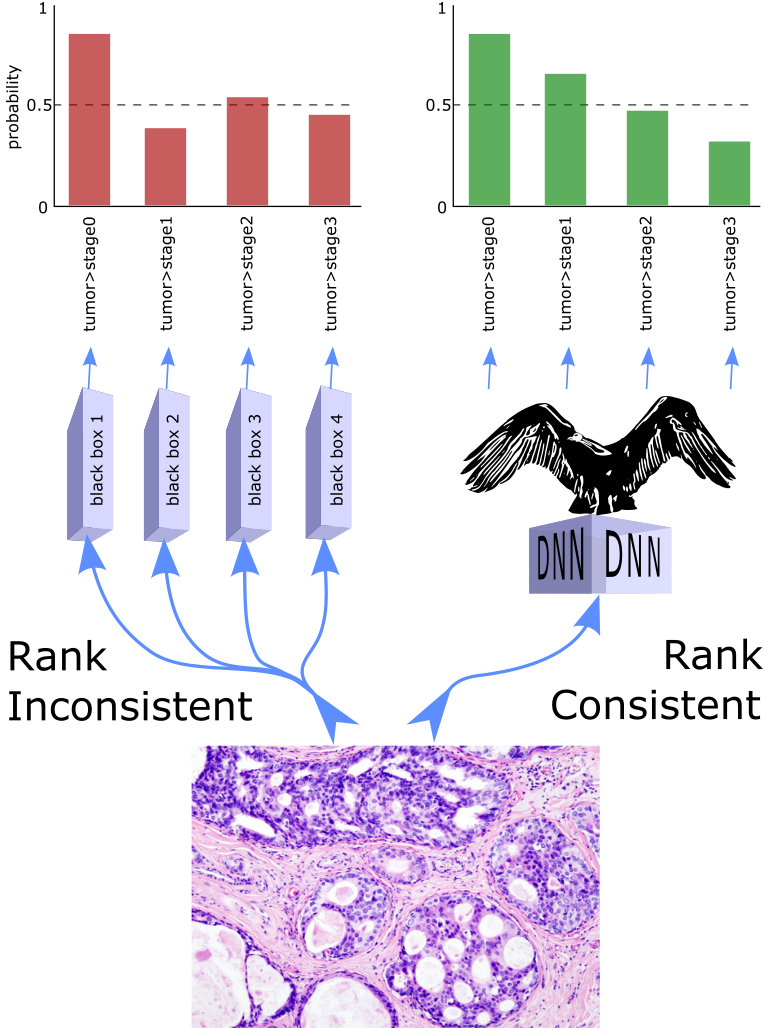}
\end{center}
\caption{Existing methods can produce rank inconsistent predictions, whereas CONDOR sits atop any DNN architecture and produces universally rank consistent results. This improves performance and interpretability of the ordinal model.}
\label{fig-1}
\end{figure}

For convenience, we often deal with the rank index $s\in\{1,\ldots,K-1\}$ associated with the rank $r_s\in\sY$. From the binary classifier subtasks, the rank index $s_i$ for input feature vector $\vx_i$ can be estimated as
\begin{equation}
s_i=1+\sum_{k=1}^{K-1} f_k(\vx_i;\vtheta), \label{eq-rankind}
\end{equation}
although multiple methods are possible to produce point estimates for $s_i$ from the probabilities $p_k(\vx_i;\vtheta),k=1,\ldots,K-1$. As shown in Figure~\ref{fig-1}, the aforementioned binary encoding approach requires that $p_k,k=1,\ldots,K-1$ be rank-monotonic
\begin{equation}
p_{1}(\vx_i;\vtheta) \geq \cdots\geq p_{K-1}(\vx_i;\vtheta), \nonumber
\end{equation}
for all $\vx_i\in\sX$ and $\vtheta\in\Theta$ to guarantee consistent predictions.

Rather than directly estimating the marginal probabilities $p_{k}(\vx;\vtheta)=P(\ry^{(k)} = 1 |\rvx=\vx,\rvtheta=\vtheta)$, $k=1,\ldots,K-1$, as in existing approaches based on
\citet{LiLin07}, 
we estimate the conditionals
\begin{eqnarray}
q_{k}(\vx;\vtheta) &:=& P(\ry^{(k)} = 1 |\rvx=\vx, \ry^{(k-1)} = 1,\ry^{(k-2)} = 1,
\ldots,\ry^{(0)} = 1,\rvtheta=\vtheta) \nonumber\\
&=& P(\ry^{(k)} = 1 |\rvx=\vx, \ry^{(k-1)} = 1,\rvtheta=\vtheta),  \label{eq-ORCAcond} 
\end{eqnarray}
for $k=1,\ldots,K-1$ where we set the boundary condition $\ry^{(0)}=1$ with unit probability by convention. Equality~(\ref{eq-ORCAcond}) follows by construction of the binary labels since $y^{(k-1)}=1$ implies $y>r_{k-1}>r_{k-2}>\cdots>r_1$, which by definition means $y^{(k')}=1$ for $k'\leq k-1$. By the same reasoning, the marginal probability is equivalent to the joint probability
\begin{equation}
p_{k}(\vx,\vtheta) ~:=~ P(\ry^{(k)}=1|\rvx=\vx,\rvtheta=\vtheta) ~=~ P(\ry^{(k)}=1,\ry^{(k-1)} = 1\ldots,\ry^{(0)}=1|\rvx=\vx,\rvtheta=\vtheta) \nonumber
\end{equation}
and by the product rule we produce a heterogeneous Markov chain representation of our marginal probabilities
\begin{equation}
p_{k}(\vx;\vtheta) = \prod_{k'=1}^{k} q_{k'}(\vx;\vtheta). 
\label{eq-ORCA}
\end{equation}

The above method in Equations (\ref{eq-ORCAcond})~\&~(\ref{eq-ORCA}) are exact (i.e., not approximations) and thus fully general, and this fact has previously been exploited in the literature for performing ordinal semantic segmentation~\citep{FernCard18}. These equations can in theory be applied in any classifier estimating binary probabilities, but we focus on the application to DNNs where $\vtheta$ represents the trainable parameters of the network. Namely, we select the final layer of the neural network to have $K-1$ nodes with sigmoid activations representing $q_{k}(\vx;\vtheta),k=1,\ldots,K-1$. For training, a reasonable heuristic loss akin to the one used in \cite{CORAL} is the weighted binary cross-entropies (WBCE) of all the subtasks
\begin{equation}
\widetilde{L}(\vtheta) = - \sum_{n=1}^N\sum_{k=1}^{K-1} \lambda_k  \Bigg(y_n^{(k)} \ln \left[\prod_{k'=1}^{k} q_{k'}(\vx_n;\vtheta) \right] +
(1-y_n^{(k)})\ln\left[1-\prod_{k'=1}^{k} q_{k'}(\vx_n;\vtheta) \right]\Bigg) 
\label{eq-condorLoss}
\end{equation}
where $\lambda_{k}>0$ is the importance parameter for task $k$, which we set to one in the subsequent. However, as we will demonstrate, minimizing the following loss function results in the maximum
likelihood (ML) estimate for the neural network parameters
\begin{equation}
L(\vtheta)=-\sum_{n=1}^N\sum_{k=1}^{K-1}y_n^{(k-1)}\Bigg( y_n^{(k)}  \ln q_k(\vx_n;\vtheta) + (1-y_n^{(k)}) \ln \left[1-q_k(\vx_n;\vtheta)\right]\Bigg).
\label{eq-MLloss}
\end{equation}
We call this approach Conditionals for Ordinal Regression (CONDOR).

\subsection{The ML loss and rank consistency of CONDOR}
Here we substantiate CONDOR's guarantee for preserving rank consistency and its ability to represent any rank consistent solution. We first provide a proof regarding the maximum likelihood loss function.

\begin{theorem}
The parameters $\vtheta$ that minimize the loss function in Equation~(\ref{eq-MLloss}) are the maximum likelihood estimators.
\label{thm-3}
\end{theorem}

\begin{proof}
Note that we have
\begin{equation}
p(\ry^{(k)}=1 | y^{(k-1)},\vx,\vtheta)=
\begin{cases}
    q_k(\vx;\vtheta),& \text{ if } y^{(k-1)}=1,\nonumber \\
    0,&\text{ if } y^{(k-1)}=0,
\end{cases}
\end{equation}
or equivalently $p(\ry^{(k)}=y^{(k)} | y^{(k-1)},\vx,\vtheta) = y^{(k)} q_k(\vx;\vtheta)y^{(k-1)} + (1-y^{(k)})(1-q_k(\vx;\vtheta)y^{(k-1)})$. And thus under independent and identically distributed data, we find the negative log likelihood $\Lambda$ of our data to be
\begin{eqnarray}
\Lambda(\vtheta)&=& -\ln \prod_{n=1}^N\prod_{k=1}^{K-1} p(\ry^{(k)}=y_n^{(k)} | y_n^{(k-1)},\vx_n,\vtheta) p(\rvx=\vx_n) \nonumber \\
&=& C-\sum_{n=1}^N\sum_{k=1}^{K-1}\ln \left[ y_n^{(k)} q_k(\vx_n;\vtheta)y_n^{(k-1)} + (1-y_n^{(k)})(1-q_k(\vx_n;\vtheta)y_n^{(k-1)}) \right] \nonumber \\
&=& C-\sum_{n=1}^N\sum_{k=1}^{K-1} y_n^{(k)}  \ln \left[q_k(\vx_n;\vtheta)y_n^{(k-1)}\right] + (1-y_n^{(k)}) \ln \left[1-q_k(\vx_n;\vtheta)y_n^{(k-1)}\right] \nonumber \\
&=& C- \sum_{n=1}^N\sum_{k=1}^{K-1}y_n^{(k-1)}\bigg( y_n^{(k)}  \ln q_k(\vx_n;\vtheta) + (1-y_n^{(k)}) \ln \left[1-q_k(\vx_n;\vtheta)\right]\bigg) =C+L(\vtheta)\nonumber 
\end{eqnarray}
where the constant $C=-\sum_{n=1}^N\ln p(\rvx=\vx_n)$ is independent of our parameters $\vtheta$ as it depends only on the (unspecified) distribution of the features, and the penultimate equality made use of the fact that \mbox{$y_n^{(k-1)}=0\implies y_n^{(k)}=0$}. Thus minimizing $L(\vtheta)$ from
Equation~(\ref{eq-MLloss}) with respect to our neural network parameters $\vtheta$ will minimize the negative
log likelihood $\Lambda(\vtheta)=C+L(\vtheta)$, and  provide the maximum likelihood estimates
of our parameters.
\end{proof}

\begin{lemma}
CONDOR provides universal rank consistency (i.e., rank consistent estimates  for all input data $\vx \in\sX$ and any parameterization $\vtheta \in\Theta$ of the DNN). 
\label{lem-1}
\end{lemma}

\begin{proof}
In neural networks, we can enforce $0 < q_{k}(\vx;\vtheta) < 1$ for all $\vx$ and any weight parameterization $\vtheta$ of the DNN by having $K-1$ output nodes  with sigmoid activations representing $q_{k}(\vx;\vtheta), k=1,\ldots,K-1 $. Because $0 < q_{k}(\vx;\vtheta) < 1$ for all $\vx$ and $\vtheta$, we have by Equation~(\ref{eq-ORCA}) 
\begin{equation}
p_{k+1}(\vx;\vtheta) = p_k(\vx;\vtheta)q_{k+1}(\vx;\vtheta) < p_k(\vx;\vtheta)\nonumber
\end{equation}
for all $\vx,\vtheta$ and $k=1,\ldots,K-2$. Thus we have rank consistency 
\begin{equation}
p_{1}(\vx;\vtheta) \geq p_{2}(\vx;\vtheta) \geq \cdots \geq p_{K-1}(\vx;\vtheta)\nonumber
\end{equation}
for all $\vx$ and any weight parameterization $\vtheta$ of the DNN.
\end{proof}

To simplify the presentation, we will suppress the notational dependence on $\vtheta$ in the remainder of the manuscript.

\begin{theorem}
Assuming that a neural network can universally approximate any $\mathcal{C}^1$ function $g:\sX\rightarrow \R^{K-1}$ such that $\widehat{g}(\vx)=g(\vx)+ \mathcal{O}(\epsilon)$ for all $\vx$ and some $\epsilon>0$, then adding a CONDOR output layer to said network can approximate any rank consistent continuous ordinal regressors $p^*_k:\sX\rightarrow [0,1],k=1,\ldots,K-1$ with error $\mathcal{O}(\epsilon)$.
\label{lem-2}
\end{theorem}
\begin{proof}
By rank consistency 
\begin{equation}
p^*_{1}(\vx_i) \geq \cdots p^*_{k-1}(\vx)\geq p^*_{k}(\vx) \cdots \geq p^*_{K-1}(\vx_i) \geq 0 \nonumber
\end{equation}
for any $\vx$  and we have defined the boundary condition $p^*_0(\vx)=1$. Then we define for $\epsilon>0$
\begin{equation}
q^*_k(\vx):=\frac{p^*_{k}(\vx)+\epsilon}{p^*_{k-1}(\vx)+2\epsilon}, ~~~~k=1,\ldots,K-1, \label{eq-q*} 
\end{equation}
and note that for each $k$ and $\vx$ the function $q_k(\vx)$ is continuous and satisfies
\begin{equation}
0< q^*_k(\vx)<1.\nonumber
\end{equation}
Then define the continuous functions
\begin{equation}
a^*_k(\vx):=\sigma^{-1}(q^*_k(\vx)) = \log \frac{q^*_k(\vx)}{1-q^*_k(\vx)}, ~~~~k=1,\ldots,K-1. \nonumber
\end{equation}
Because the upstream neural network can approximate any continuous function $g:\sX\rightarrow \R^{K-1}$ with error $\mathcal{O}(\epsilon)$, we can set $g_k(\vx)=a_k^*(\vx)$ and have the upstream neural network produce the functions $\widehat{g}_k(\vx)=a_k^*(\vx)+\mathcal{O}(\epsilon)$ for all $k$ and $\vx$. Then after the CONDOR sigmoid activations we would have the neural network produce for all $\vx$ at its output nodes\footnote{Note that only the activation at the last layer is fixed as the sigmoid function. The activation of the hidden layers could be other functions, e.g., ReLU.}
\begin{equation}
\sigma(a^*_k(\vx)+\mathcal{O}(\epsilon)), ~~~~k=1,\ldots,K-1, \nonumber
\end{equation}
which by Taylor Series about $a^*_k(\vx)$ produces
\begin{eqnarray}
\sigma(a^*_k(\vx)+\mathcal{O}(\epsilon))&=& \sigma(a^*_k(\vx))+ \sigma'(a^*_k(\vx))\mathcal{O}(\epsilon) + \sigma''(a^*_k(\vx))\mathcal{O}(\epsilon^2)/2 +\ldots \nonumber\\
&=&\sigma(a^*_k(\vx))+ \sigma(a^*_k(\vx))(1-\sigma(a^*_k(\vx)))\mathcal{O}(\epsilon) + \mathcal{O}(\epsilon^2)\nonumber\\
&=&\sigma(a^*_k(\vx))+ \mathcal{O}(\epsilon)\nonumber\\
&=&q^*_k(\vx)+ \mathcal{O}(\epsilon).\nonumber
\end{eqnarray}
Then the CONDOR approach yields 
\begin{eqnarray}
p_{k}(\vx)&=&\prod_{k'=1}^{k}\sigma(a^*_k(\vx)+\mathcal{O}(\epsilon)) = \prod_{k'=1}^{k} \left[q^*_{k'}(\vx)+\mathcal{O}(\epsilon)\right]\nonumber\\
&=&\mathcal{O}(\epsilon)+q^*_{1}(\vx)\prod_{k'=2}^{k} \left[q^*_{k'}(\vx)+\mathcal{O}(\epsilon)\right]\nonumber\\
&=&\mathcal{O}(\epsilon)+q^*_{1}(\vx)q^*_{2}(\vx)\prod_{k'=3}^{k} \left[q^*_{k'}(\vx)+\mathcal{O}(\epsilon)\right].\nonumber
\end{eqnarray}
By iteration it follows that
\begin{align*}
    p_{k}(\vx)&= \left[\prod_{k'=1}^{k} q^*_{k'}(\vx)\right] +\mathcal{O}(\epsilon).
\end{align*}
Using the definition of $q^*_k(\cdot)$ in
Equation~(\ref{eq-q*}), we get
\begin{eqnarray}
p_{k}(\vx)&=&\left[ \frac{\prod_{k'=1}^{k}[p^*_{k'}(\vx)+\epsilon]}{\prod_{k'=1}^{k}[p^*_{k'-1}(\vx)+2\epsilon]}\right] +\mathcal{O}(\epsilon) \nonumber\\
&=&\frac{[\prod_{k'=1}^{k}p^*_{k'}(\vx)]+\mathcal{O}(\epsilon)}{[\prod_{k'=1}^{k}p^*_{k'-1}(\vx)]+\mathcal{O}(\epsilon)} +\mathcal{O}(\epsilon) \nonumber\\
&=&\frac{\prod_{k'=1}^{k}p^*_{k'}(\vx)}{[\prod_{k'=1}^{k}p^*_{k'-1}(\vx)]+\mathcal{O}(\epsilon)} +\frac{\mathcal{O}(\epsilon)}{[\prod_{k'=1}^{k}p^*_{k'-1}(\vx)]+\mathcal{O}(\epsilon)}+\mathcal{O}(\epsilon) \nonumber\\
&=&\frac{\prod_{k'=1}^{k}p^*_{k'}(\vx)}{[\prod_{k'=1}^{k}p^*_{k'-1}(\vx)]+\mathcal{O}(\epsilon)} +\mathcal{O}(\epsilon) \label{eq-semifinal} \\
&=& p^*_{k}(\vx)+\mathcal{O}(\epsilon) \label{eq-final}
\end{eqnarray}
for all $\vx$ and $k=1,\ldots,K-1$. The last Equality~(\ref{eq-final}) comes from considering separately the cases $\prod_{k'=1}^{k}p^*_{k'-1}(\vx)=0$ and $\prod_{k'=1}^{k}p^*_{k'-1}(\vx)>0$. In the former we have $p_k^*(\vx)=0$ and Equation~(\ref{eq-semifinal}) reduces to $p_k(\vx)=0+\mathcal{O}(\epsilon)$. In the latter we find from Equation~(\ref{eq-semifinal})
\begin{eqnarray}
p_{k}(\vx)
&=&\frac{1/\prod_{k'=1}^{k}p^*_{k'-1}(\vx)}{1/\prod_{k'=1}^{k}p^*_{k'-1}(\vx)}\times \frac{[\prod_{k'=1}^{k}p^*_{k'}(\vx)]}{[\prod_{k'=1}^{k}p^*_{k'-1}(\vx)+\mathcal{O}(\epsilon)]} +\mathcal{O}(\epsilon) \nonumber \\
&=&\frac{p^*_{k}(\vx)}{[1+\mathcal{O}(\epsilon)]} +\mathcal{O}(\epsilon) \nonumber\\
&=&p^*_{k}(\vx)[1+\mathcal{O}(\epsilon)] 
+\mathcal{O}(\epsilon) \label{eq-powerseries}\\
&=& p^*_{k}(\vx)+\mathcal{O}(\epsilon) \nonumber
\end{eqnarray}
where Equality~(\ref{eq-powerseries}) comes from the
power series $1/(1+x)=1-x+x^2-\ldots$.
\end{proof}

\section{Numerical Experiments}
In the subsequent sections, we then demonstrate CONDOR's superior  performance compared to the state-of-the-art on several benchmark and real-world data sets.

Specifically, we profile the WBCE from Equation~(\ref{eq-condorLoss}), the earth movers distance (EMD) on the rank indices (assuming unit distance between ranks), and the mean absolute error (MAE) on the rank indices (assuming unit distance between ranks)  using Equation~(\ref{eq-rankind}) for the point estimate. We  benchmark CONDOR trained using the maximum-likelihood loss in Equation~(\ref{eq-MLloss}) as well as CONDOR trained using the WBCE loss in Equation~(\ref{eq-condorLoss}), which will be denoted as CONDOR-WBCE in the subsequent. 

In these experiments, the only difference between the four methods is the choice of loss function and final layer of the neural network; all other details of the DNN architecture, the optimization algorithms,  hyperparameters and random number seeds are kept equal throughout each experiment. Namely, CORAL and CONDOR-WBCE both have the WBCE for the $K-1$ subtasks as a loss, whereas Categorical uses the CCE and CONDOR uses Equation~(\ref{eq-MLloss}). Likewise, CORAL uses a custom final layer with weight sharing~\cite{CORAL} among $K-1$ output nodes, CONDOR uses a final dense layer with $K-1$ output nodes which after sigmoid activation represent $q_k(\vx),k=1,\ldots,K-1$, and the categorical algorithm uses a dense layer with $K$ nodes and a softmax activation. All results were gathered with three random number seeds and reported as the mean plus or minus the standard deviation across these seeds.

\subsection{Synthetic Quadrants Dataset}
We consider the simple task of ordinal classification wherein the labels $0,1,2,3$ are the quadrants of the plane going counterclockwise and the features are generated from a 2D standard normal distribution. We draw $1000$ samples and do a $90/10$ train/test split of the dataset. We select as the upstream network architecture two dense layers with ten neurons and RELU activations and an Adam optimizer with $100$ epochs and early stopping patience of $10$ using a validation split of $0.2$. As can be seen in Table~\ref{syth-table}, the algorithms proposed in CONDOR demonstrate the best performance in WBCE, EMD and MAE.

\begin{table}[h]
\caption{Synthetic quadrants data results in test set}
\label{syth-table}
\begin{center}
\begin{tabular}{lccc}
\multicolumn{1}{c}{\bf ALGORITHM} &\multicolumn{1}{c}{\bf WBCE} &\multicolumn{1}{c}{\bf MAE}&\multicolumn{1}{c}{\bf EMD}

\\ \hline \\
CONDOR &  {0.1113 $\pm$ 0.0615} &{\bf 0.0033 $\pm$ 0.0047}& {0.0955 $\pm$ 0.0526}  \\
CONDOR-WBCE &  {\bf 0.0768 $\pm$ 0.0100} &{0.0167 $\pm$ 0.0153}& {\bf 0.0799 $\pm$ 0.0526}  \\
CORAL  &0.5074 $\pm$ 0.0754 &0.0733 $\pm$ 0.0416 &  0.4080 $\pm$ 0.0377  \\
CATEGORICAL & 1.3438 $\pm$ 0.0458 &  0.0200 $\pm$ 0.0173  & 1.0318 $\pm$ 0.0267  \\
\end{tabular}
\end{center}
\end{table}

\subsection{MNIST as an ordinal dataset}
Depending on the application, MNIST can be considered a categorical problem or an interval regression problem. If the digits are used for license-plate recognition, then the problem is categorical since there is no notion of ``close'' errors. By contrast, if the digits are used for GPS coordinates or postal codes, then the ordering and distance between numerals becomes relevant and categorical classification is no longer the most appropriate framing of the task. It is valid to treat interval regression as an ordinal problem since this assumes less structure on the dataset, although it is recommended to exploit the interval scale. Here we treat MNIST as ordinal data for the purpose of benchmarking our ordinal algorithms, while acknowledging that it should likely be treated as either a categorical classification or interval regression as dictated by the specific real-world application setting. The MNIST data are split into training, validation and test sets of 55K, 5K and 10K images, respectively.  We utilize a convolutional neural network with two convolutional layers of 64 and 32 filters respectively and a kernel size of 3, before flattening and passing to the appropriate output layer and loss function for our four models (CONDOR, CONDOR-WBCE, CORAL, Categorical).  Training is performed with the Adam optimizer, a maximum of $100$ epochs and an early-stopping patience of $10$. The results in Table~\ref{mnist-table} indicate that CONDOR demonstrates superior performance in all three metrics.  

\begin{table}[h!]
\caption{MNIST digit image classification on test set}
\label{mnist-table}
\begin{center}
\begin{tabular}{lccc}
\multicolumn{1}{c}{\bf ALGORITHM} &\multicolumn{1}{c}{\bf WBCE} &\multicolumn{1}{c}{\bf MAE}&\multicolumn{1}{c}{\bf EMD}
\\ \hline \\
CONDOR &  {\bf 0.1721
 $\pm$ 0.0091
} &{\bf 0.0589
 $\pm$ 0.0053
}& {\bf 0.0808
 $\pm$ 0.0045
}  \\
CONDOR-WBCE &  {0.1784
 $\pm$ 0.0043
} &{ 0.0596
 $\pm$ 0.0027
}& {0.0818
 $\pm$ 0.0065
}  \\
CORAL  &1.2724
 $\pm$ 0.0139
 &0.4583
 $\pm$  0.0028
 & 0.7501
 $\pm$  0.0214
 \\
CATEGORICAL & 5.5424
 $\pm$ 0.0013
 & {0.0592
 $\pm$ 0.0042}
  & 3.0638
 $\pm$  0.0013
 \\
\end{tabular}
\end{center}
\end{table}

\subsection{NLP on Amazon reviews dataset}
Here we consider a natural language processing (NLP) dataset consisting of $99,025$ (non-duplicate and non-empty) Amazon Pantry text reviews with their corresponding one to five star ratings~\citep{amazon}. We split the data to have a test set with $10,000$ examples. For the neural network architecture, we use the fixed and pre-trained Google universal sentence encoder~\citep{Cer:18} and append a dense layer with 64 ReLu-activated neurons and a dropout of $0.1$, followed by the appropriate output layer and loss function for each model. Training is performed with the Adam optimizer, a maximum of $100$ epochs and an early-stopping patience of 10 with a validation split of $0.2$. The results in Table~\ref{amazon-table} demonstrate that CONDOR-WBCE provides the strongest performance in this benchmark across all three performance metrics.

\begin{table}[h]
\caption{Amazon product review text classification on test set}
\label{amazon-table}
\begin{center}
\begin{tabular}{lccc}
\multicolumn{1}{c}{\bf ALGORITHM} &\multicolumn{1}{c}{\bf WBCE} &\multicolumn{1}{c}{\bf MAE}&\multicolumn{1}{c}{\bf EMD}
\\ \hline \\
CONDOR &  {0.7816
 $\pm$ 0.0125
} & {0.3185
 $\pm$ 0.0049
} & {0.4641
 $\pm$ 0.0075
}  \\
CONDOR-WBCE &  {\bf 0.7807
 $\pm$ 0.0113
} & {\bf 0.3180
 $\pm$ 0.0047
} & {\bf 0.4582
 $\pm$ 0.0050
}  \\
CORAL  &0.8095
 $\pm$ 0.0098
 &0.3263
 $\pm$  0.0052
 & 0.4726
 $\pm$  0.0041
 \\
CATEGORICAL & 2.4663
 $\pm$ 0.4678
 & 0.4195
 $\pm$ 0.0074
  & 1.6906
 $\pm$ 0.2422
   \\
\end{tabular}
\end{center}
\end{table}

\subsection{GRU-D for COVID-19 prognostication}
This study adheres to a research protocol approved by the Mayo Clinic Institutional Review Board. Here we extend the results from \citet{grud21} to progress from their binary classification predicting mortality to ordinal regression predicting severity of outcomes. Namely, this clinical dataset includes two binary severity outcomes: an indicator variable for mechanical ventilation or extracorporeal membrane oxygenation (ECMO), as well as an indicator variable of whether patient death occurred. From these, a clear three point ordinal scale can be constructed whereby a patient is scored a zero when they have no severe outcome, a one when they experienced the severe outcome of ventilation or ECMO, and a two corresponding to death (with or without prior ventilation or ECMO). \citet{grud21} identified the GRU-D recurrent neural network architecture as the best performing model for binary mortality prediction, and we extend that approach to the ordinal problem. We do this for the CONDOR, CORAL, and categorical algorithms using their corresponding final layers and loss functions. This GRU-D architecture~\citep{Che:18} deals explicitly with the $55$ dimensional time series that is missing not at random (MNAR) due to the manner in which clinical data is ordered and recorded in an electronic health record (EHR). The default hyperparameters (dropout of $0.3$, l2 regularizaton of $0.0001$, $100$ hidden and recurrent neurons, batch size of $256$, adam learning rate $0.001$, no batch norm, no bidirectional RNN, $50$ max time steps, $100$ epochs with early stopping patience of $10$ epochs) have previously demonstrated strong performance ~\citep{grud21} and so are retained here. 

\begin{table}[ht]
\caption{COVID-19 severity prediction results in prospective test set}
\label{covid-table}
\begin{center}
\begin{tabular}{lccc}
\multicolumn{1}{c}{\bf ALGORITHM} &\multicolumn{1}{c}{\bf WBCE} &\multicolumn{1}{c}{\bf MAE}&\multicolumn{1}{c}{\bf EMD}
\\ \hline \\
CONDOR &  {0.6572 $\pm$ 0.0036} &{0.2993 $\pm$ 0.0040}& {\bf 0.4144 $\pm$ 0.0073}   \\
CONDOR-WBCE &  {\bf 0.6526 $\pm$ 0.0004} &{\bf 0.2986 $\pm$ 0.0044}& {0.4151 $\pm$ 0.0063}   \\
CORAL  &0.6711 $\pm$ 0.0000 & 0.3021 $\pm$ 0.0015 & 0.4261 $\pm$ 0.0023  \\
CATEGORICAL & 1.1075 $\pm$ 0.0019 & 0.3076 $\pm$ 0.0021  & 0.8185 $\pm$ 0.0010  \\
\end{tabular}
\end{center}
\end{table}

The dataset is split into a training/validation set of $9,435$ patients who tested positive for COVID-19 prior to December 15 2020 by PCR test, and a prospective testing set of $2,372$ patients who tested positive on or after that date. For training  we use an identical $90/10$ training/validation split on the $9,435$, which facilitates early stopping with patience.  The results in Table~\ref{covid-table} demonstrate that CONDOR is superior in EMD while CONDOR-WBCE is superior in the remaining metrics. Furthermore, the CONDOR-WBCE GRU-D model has a prospective test set AUROC of $0.9038 \pm 0.0025$ for the mortality prediction subtask, which is greater than the 0.901 reported in \citet{grud21} wherein the authors trained the algorithm as a binary classifier specifically for mortality prediction. This demonstrates that there is no loss of mortality prediction performance when building a DNN to address the more challenging task of prognostication.

\section{Discussion}
We have demonstrated the ability of the CONDOR approach to overcome limitations present in popular alternative methods and to produce rank consistent results in the classification of data with ordinal labels. Rank consistency is not only important for theoretical soundness, but in application settings where explainability is important and a rank inconsistent prediction will be unacceptably contradictory and fundamentally unexplainable. Regardless of the loss function being optimized or the parameterization of the neural network, CONDOR provides universal guarantees of rank consistency by Lemma~\ref{lem-1}, which is to say the CONDOR approach is ``sufficient'' for rank consistency. Our next result leverages the fact that there are a wide-variety of universal approximation theorems for neural networks each with their own technical conditions (e.g., see \citet{Chong20} for discussion of various universal approximation proofs and technical conditions). Namely, Theorem~\ref{lem-2} states than any upstream neural network satisfying the conditions for universal approximation can be provided a CONDOR output layer, which will create a universally rank consistent network that can approximate any rank consistent solution. This theorem can be interpreted as CONDOR being ``necessary'' for rank consistency, insofar as any rank consistent solution can be represented by a CONDOR neural network. Finally, we provide in Theorem~\ref{thm-3} the maximum likelihood loss function, which should also be more numerically stable than the heuristic loss of WBCE particularly when the number of ordinal classes is large. Thus we suggest the maximum likelihood loss in practice, even though its performance was similar to WBCE in our numerical experiments. 

In contrast to our theoretical results, note that Theorem~1 of \citet{CORAL} only provides rank consistency at the global minimum of an optimization problem with the specified loss function. Since neural network training is not guaranteed or expected to reach a  global optimal parameterization, the \citet{CORAL} approach can in theory produce rank inconsistent solutions and thus in practice requires \emph{post hoc} checks of the estimated bias terms to verify rank consistency.  Furthermore, \citet{CORAL} restricts expressiveness in the $K-1$ binary classifier outputs that must have ``parallel slopes'' (i.e., differ only by a bias parameter whose impact is completely independent of the feature vector). In Appendix~\ref{sec-CORALproof}, we formalize these comments with two proofs demonstrating that the CORAL framework~\citep{CORAL} lacks the theoretical guarantees of CONDOR. 

Beyond our mathematical justifications, ultimately it is critical that the method perform well within a wide variety of neural network architectures and ordinal problem settings. Our benchmarking of dense networks, CNNs, attention networks, and exotic RNNs all shows practical benefit of using the CONDOR algorithm in a diverse set of ordinal applications using a variety of ordinal metrics. Furthermore, beyond ordinal measures of performance, CONDOR remains competitive in the categorical performance measure of accuracy, and in-fact provides improved classification in true ordinal problems when compared to categorically optimized neural networks. 
We attribute this to the ability of the network to exploit ``clues"  encountered during ordinal training (Appendix~\ref{sec-acc}).

In addition to the theoretical strengths and performance improvements of the CONDOR method, we note that many applied machine learning papers simply use categorical classification in their problem settings rather than consider current state-of-the-art methods for ordinal regression. Part of this may be educational, as most beginners are only taught binary/categorical classification and continuous regression. However, the authors also believe part of the barrier is programmatic ease-of-use. We provide production-ready and user-friendly software packages in both PyTorch and TensorFlow, in order to minimize the effort required to convert existing categorical code-bases into CONDOR ordinal code-bases. In Appendix~\ref{sec-code}, we demonstrate the modest code changes required to implement our methodology in an existing categorical code base.

The key requirements for successful supervised learning tasks are algorithms that respect the structure of the problem, and access to sufficient amounts of labeled data. Since CONDOR satisfies the first requirement by providing a robust algorithm for ordinal regression, we conclude with the latter by emphasizing the prevalence of available ordinal outcome measurements, using  medical applications as a prototypical applied problem domain. Survey research for instance, frequently utilizes ordinal responses such as the psychometric Likert scale~\citep{Lik32}, providing a large corpus of existing data with ordinal labels. Furthermore, while labeling outcomes from the Electronic Health Record (EHR) is one of the most time-consuming and expensive aspects of applied machine learning in the medical space, the proliferation of ordinal scales in modern medical practice (see Appendix~\ref{sec-ordMed}) means the EHR already contains physician-provided ordinal outcomes from a large variety of settings. The ubiquity of ordinal outcome measurements throughout survey research and medical settings represents a rich untapped reserve of training data that have yet to be fully explored by ordinal regression machine learning algorithms, and it is our hope that CONDOR's demonstrated capabilities and its ease-of-use will encourage its adoption and enable broad exploration of underutilized data across these and other domains.

\subsubsection*{Acknowledgments}
The funding for this research has been provided Mayo Clinic Center for Individualized Medicine. The funders had no role in study design, data collection and analysis, decision to publish, or preparation of the manuscript. The authors thank Saranya Sankaranarayanan and Jagadheshwar Balan for sharing their preprocessed versions of the COVID-19 data set and  
their code for GRU-D mortality prediction.

\bibliographystyle{iclr2022_conference}
\bibliography{references}

\appendix
\section{Appendices}

\subsection{CORAL proofs} \label{sec-CORALproof}
In this appendix, we demonstrate formally that the CORAL framework of \citet{CORAL} does not have the universal rank consistency nor the expressiveness of CONDOR.
\begin{lemma}
CORAL is not universally rank consistent.
\end{lemma}
\begin{proof}
For CORAL the last layer shares weights and only has different bias terms $b_1,\ldots,b_k$ meaning it represents the output probabilities as~\citep{CORAL}
\begin{equation}
p_k(\vx)=\sigma(a(\vx)+b_k)=\frac{1}{1+\exp(-a(\vx))\exp(-b_k)}~~~~k=1,\ldots,K-1, \nonumber
\end{equation}
for some $a(\vx)$. This means in the notation of CONDOR that
\begin{eqnarray}
q_k(\vx)&=&\frac{p_{k}(\vx)}{p_{k-1}(\vx)}=
\frac{1+\exp(-a(\vx))\exp(-b_{k-1})}{1+\exp(-a(\vx))\exp(-b_k)} \nonumber \\
&=&\frac{\exp(a(\vx))+\exp(-b_{k-1})}{\exp(a(\vx))+\exp(-b_k)} \label{eq-coral}
\end{eqnarray}
for $k=2,\ldots,K-1$ and $q_1(\vx)=p_1(\vx)$. Note that if the neural network parameters are chosen such that $b_k>b_{k-1}$ for some $k$ then $q_k(\vx)>1$ and CORAL is rank inconsistent.
\end{proof}

It is quite clear from Equation~(\ref{eq-coral}) that the functional form of CORAL is far more restrictive than CONDOR, which allows arbitrary $q_k:\sX\rightarrow [0,1]$ functions. For completeness, we prove formally in the next lemma that it is less expressive.

\begin{lemma}
CORAL can not approximate every rank consistent solution with $\mathcal{O}(\epsilon)$ error.
\end{lemma}
\begin{proof}
For simplicity, consider a univariate input $x$ and $K=3$. Then we have for CORAL
\begin{eqnarray}
q_2(x)&=&\frac{p_{2}(x)}{p_{1}(x)}= \frac{1+\exp(-a(x))\exp(-b_{1})}{1+\exp(-a(x))\exp(-b_2)} \nonumber \\
&=&\frac{\exp(a(x))+\exp(-b_{1})}{\exp(a(x))+\exp(-b_2)}, \nonumber
\end{eqnarray}
and $q_1(x)=p_1(x)=\sigma(a(x)+b_1)$. Consider an extremely simple CONDOR network with no hidden layers, bias parameters fixed to zero and only two weights $w_1=1,w_2=2$ producing
\begin{eqnarray}
q^*_k(x)&=& \frac{1}{1+\exp(kx)},~~~~k=1,2 \nonumber
\end{eqnarray}
which is rank consistent by Lemma~\ref{lem-1}. Suppose by way of contradiction that there exists CORAL $a(x)$ and $b_1,b_2$ such that $q_k(x)=q_k^*(x)+\mathcal{O}(\epsilon)$ for all $k$ and $x$. Thus
\begin{eqnarray}
\frac{1}{1+\exp(-a(x))\exp(-b_1)}&=& \frac{1}{1+\exp(x)} +\mathcal{O}(\epsilon) \nonumber \\
\frac{\exp(a(x))+\exp(-b_1)}{\exp(a(x))+\exp(-b_2)}&=& \frac{1}{1+\exp(2x)}+\mathcal{O}(\epsilon) \nonumber 
\end{eqnarray}
which implies
\begin{eqnarray}
\exp (a(x))&=&\exp(-x-b_1) +\mathcal{O}(\epsilon) \label{eq-ax} \\
(\exp(a(x))+\exp(-b_1))(1+\exp(2x))&=& \exp(a(x))+\exp(-b_2) +\mathcal{O}(\epsilon) \label{eq-ax1} 
\end{eqnarray}
and plugging in Equation~(\ref{eq-ax}) into Equation~(\ref{eq-ax1}) we find after rearrangement
\begin{eqnarray}
\exp(x)+\exp(2x)&=& \exp(b_1-b_2)-1 +\mathcal{O}(\epsilon)\nonumber 
\end{eqnarray}
which is a contradiction since the left hand side has an infinite range depending on $x$ whereas the right hand side is a constant up to an error of order $\epsilon$.
\end{proof}

\subsection{Categorical accuracy results}\label{sec-acc}
Accuracy is a categorical performance measure wherein there is not an increasing penalty for being further from the correct label, and therefore no relative ``credit'' given for being close to the correct label. One might therefore expect that training a neural network with a categorical loss (i.e., CCE) would result in higher categorical accuracy than if the network were trained with a ordinal method. 

\begin{table}[ht]
\caption{Accuracy in benchmark test sets}
\label{tab-acc}
\begin{center}
\begin{tabular}{lcccc}
\multicolumn{1}{c}{\bf ALGORITHM} &\multicolumn{1}{c}{\bf Quadrants} &\multicolumn{1}{c}{\bf MNIST}&\multicolumn{1}{c}{\bf Amazon}&\multicolumn{1}{c}{\bf COVID-19}
\\ \hline \\
CONDOR & \bf 0.9967 $\pm$ 0.0047 & 0.9834$\pm$ 0.0008  & {0.7493 $\pm$ 0.0027 }& {\bf 0.7250 $\pm$ 0.0031}  \\
CONDOR-WBCE & 0.9900 $\pm$ 0.0100 & 0.9805 $\pm$ 0.0004  & {\bf 0.7498 $\pm$ 0.0032 }& {\bf 0.7250 $\pm$ 0.0033}  \\
CORAL  & 0.9333 $\pm$ 0.0379 &  0.6381 $\pm$ 0.0037 & 0.7343 $\pm$ 0.0036 & 0.7175 $\pm$ 0.0015 \\
CATEGORICAL & {0.9933 $\pm$ 0.0058} & {\bf 0.9845 $\pm$ 0.0008 }&  0.7299
 $\pm$ 0.0037
& 0.7244 $\pm$ 0.0016 \\
\end{tabular}
\end{center}
\end{table}

Surprisingly, in Table~\ref{tab-acc} we find in the majority of benchmarks the proposed ordinal methods provided higher categorical accuracy than the networks trained using categorical cross entropy to specifically optimize categorical performance. We attribute this remarkable finding to the ``clues'' provided to the network when the ordinal nature of the problem is exploited during training. For instance, if the network incorrectly guesses rank index 7 when the true rank index is 8, the categorical loss treats this equivalently to a guess of a rank index of 1; the back propagation does not send any signals indicating that the guess of rank 7 was ``close'' to the true rank of 8. By contrast, Equation~(\ref{eq-condorLoss}) would capture the fact that most of the binary subtasks are correctly predicted when a rank 7 is estimated for a ground truth rank of 8. In problems like MNIST where the features are not necessarily trending with increasing rank, we could understand how a categorical loss produces a stronger categorical accuracy. But in problems like Amazon star ratings, where the language and sentiment of 4 and 5 star reviews are likely closer in the NLP embedding space than the language and sentiment of 1 star reviews, one can also understand how training with an ordinal loss could provide higher categorical accuracy than a categorical loss that only has the capacity to indicate ``correct'' versus ``incorrect'' and never ``incorrect but close''. 

\subsection{Minimal code changes}
\label{sec-code}

\begin{figure}[ht]
\begin{center}
\includegraphics[width=\linewidth]{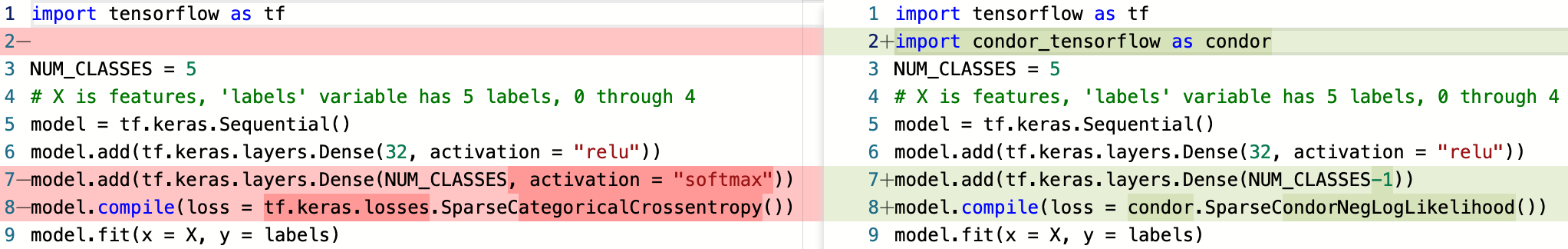}
\end{center}
\caption{Converting a categorical code base into a CONDOR ordinal code base is straightforward.}
\label{fig-codediff}
\end{figure}

CONDOR is open sourced as both TensorFlow\footnote{\url{https://github.com/GarrettJenkinson/condor_tensorflow}} and PyTorch\footnote{\url{https://github.com/GarrettJenkinson/condor_pytorch}} repositories that make it simple to modifying existing categorical code bases to use CONDOR. See Figure~\ref{fig-codediff} for a hypothetical example in TensorFlow. Both the TensorFlow and PyTorch versions of the GitHub repositories have full mkdocs documentation, docker files, ipynb tutorials, continuous integration testing and pip packaging. The authors hope this reduces the barrier to using proper and cutting-edge ordinal regression in applied problems. 

\subsection{Ordinal outcomes in medical practice}
\label{sec-ordMed}

Modern medical practice emphasizes the importance of a standardized and reproducible communication of findings and outcomes within the global medical community. Consistent diagnosis, prognostication, treatment, and decision making all require evidence- and consensus-based labeling of patient disease states. Frequently, these categorizations are made ordinal to align with the expected prognosis or severity of disease. 

As a result, across nearly every sub-specialty of medicine, one can find a plethora of ordinal outcome scales. Well-known to the general public is the use of tumor staging in oncology to characterize neoplasms. However, we provide a non-comprehensive sampling of other specialties that are perhaps less well-known. For instance, the American Association for the Surgery of Trauma provides 32 ordinal scales~\citep{AAST10} for assessing the severity of trauma to 32 organs on scale of 1 (minimal) to 6 (lethal). Molecular testing results, such as DNA variant sequencing, are often graded on an ordinal scale such as The American College of Medical Genetics and Genomics' variant scoring from variants from benign, likely benign, variant of unknown significance, likely pathogenic, to pathogenic~\citep{ACMG15}. Subjective outcomes are often measured on ordinal scales, such as the 11-level  Pain Rating Scale~\citep{pain05} from 0 (no pain) to 10 (worst possible pain). Traumatic brain injury is assessed on the 5-point Glasgow outcome scale~\citep{tbi17}. Radiologists make frequent use of ordinal scales, including but not limited to Lung-RADS screening lesions from 0 to 4~\citep{lungrads18}, BI-RADS for breast cancer screening from 0 to 6~\citep{birads14}, and Bozniak classification for renal lesions from 1 to 4~\citep{bosniak19}. Retinopathy has the Scottish Grading protocol from 0 to 4~\citep{ScottishGP15}. While these examples are not intended to be a comprehensive review, they hopefully provide some insight into just how prevalent ordinal scales are in modern medicine.

\end{document}